\documentclass[sn-mathphys,Numbered]{sn-jnl}

\usepackage{graphicx}%
\usepackage{multirow}%
\usepackage{amsmath,amssymb,amsfonts}%
\usepackage{amsthm}%
\usepackage{mathrsfs}%
\usepackage[title]{appendix}%
\usepackage{xcolor}%
\usepackage{textcomp}%
\usepackage{manyfoot}%
\usepackage{booktabs}%
\usepackage{algorithm}%
\usepackage{algorithmicx}%
\usepackage{algpseudocode}%
\usepackage{listings}%

\theoremstyle{thmstyleone}%
\theoremstyle{thmstyletwo}%

\theoremstyle{thmstylethree}%

\begin{document}

\title[Article Title]{Empowering Agricultural Insights: RiceLeafBD - A Novel Dataset and Optimal Model Selection for Rice Leaf Disease Diagnosis through Transfer Learning Technique}


\author[2]{\fnm{Sadia Afrin} \sur{Rimi}}\email{sadia.afrin.rimi.swe@gmail.com}
\equalcont{These authors contributed equally to this work.}

\author*[1]{\fnm{Md. Jalal Uddin} \sur{Chowdhury}}\email{jalal\_cse@lus.ac.bd}
\equalcont{These authors contributed equally to this work.}

\author[2]{\fnm{Rifat} \sur{Abdullah}}\email{rifat.abdullahprn@gmail.com}
\equalcont{These authors contributed equally to this work.}

\author[1]{\fnm{Iftekhar} \sur{Ahmed}}\email{iftekharifat007@gmail.com}
\equalcont{These authors contributed equally to this work.}

\author[3]{\fnm{Mahrima Akter} \sur{Mim}}\email{mahrimamim17@gmail.com}
\equalcont{These authors contributed equally to this work.}

\author[1]{\fnm{Mohammad Shoaib} \sur{Rahman}}\email{shoaib\_cse@lus.ac.bd}
\equalcont{These authors contributed equally to this work.}

\affil[1]{\orgdiv{Department of Computer Science and Engineering}, \orgname{Leading University},\city{Sylhet}, \country{Bangladesh}}

\affil[2]{\orgdiv{Department of Software Engineering}, \orgname{Daffodil International University}, \city{Dhaka}, \country{Bangladesh}}

\affil[3]{\orgdiv{Department of Business}, \orgname{Queensborough Community College}, \city{Bayside, Queens}, \country{New York}}



\abstract{The number of people living in this agricultural nation of ours, which is surrounded by lush greenery, is growing on a daily basis. As a result of this, the level of arable land is decreasing, as well as residential houses and industrial factories. The food crisis is becoming the main threat for us in the upcoming days. Because on the one hand, the population is increasing, and on the other hand, the amount of food crop production is decreasing due to the attack of diseases. Rice is one of the most significant cultivated crops since it provides food for more than half of the world's population. Bangladesh is dependent on rice (Oryza sativa) as a vital crop for its agriculture, but it faces a significant problem as a result of the ongoing decline in rice yield brought on by common diseases. Early disease detection is the main difficulty in rice crop cultivation. In this paper, we proposed our own dataset, which was collected from the Bangladesh field, and also applied Deep learning and Transfer learning models for the evaluation of the datasets. We elaborately explain our dataset and also give direction for further research work to serve society using this dataset. We applied a light CNN model and pre-trained InceptionNet-V2, EfficientNet-V2, and MobileNet-V2 models, which achieved 91.5\% performance for the EfficientNet-V2 model of this work. The results obtained assaulted other models and even exceeded approaches that are considered to be part of the state of the art. It has been demonstrated by this study that it is possible to precisely and effectively identify diseases that affect rice leaves using this unbiased datasets. After analysis of the performance of different models, the proposed datasets are significant for the society for research work to provide solutions for decreasing rice leaf disease.}




\maketitle

\section{Introduction}\label{sec1}

As the backbone of economic stability, agriculture is of utmost importance, especially in countries such as Bangladesh, where it is the primary contributor to GDP. An essential food crop, rice farming is critical to maintaining economic stability and creating significant job opportunities\cite{in1}. Nonetheless, the agricultural industry has significant obstacles, chief among which is the enduring danger of illnesses, which significantly jeopardizes rice farming and influences the amount and quality of output\cite{in2}. Traditionally, disease diagnosis techniques that rely on subjective visual cues or time-consuming laboratory testing highlight subjectivity, deadlines, and a reliance on expert knowledge\cite{in3}. In light of these constraints, modern agricultural research is gradually incorporating cutting-edge technology with an emphasis on the revolutionary potential of computer vision and machine learning to improve and automate disease identification procedures\cite{in4}.

In the last few decades, a lot of work has gone into collecting data. Several sets of data have been acknowledged. PlantVillage \cite{in6}, which has 54,000 images of the underside side of leaves on a plain background, is the most well-known in this area. But, as has been seen in other studies \cite{in7}, these combinations fail to demonstrate that the goals of the end program are adequate. The datasets that were made under controlled conditions—that is, with the leaf on a background that is all the same—do not accurately reflect the real-world conditions in which the model will work.

The importance of this study has been created clear on two levels in this circumstance. We add a new dataset to the literature called RiceLeafBD that can be used to diagnose and keep an eye on plant constraints. It consists of a set of 1555 images taken in real-life fields that show four types of leaf stress: Leaf Blight, Tungro virus, Brown Spot, and healthy leaves. We are doing a study of public images that were made to help classify and identify leaf diseases. We focus on datasets that are available on sites for sharing data in an open way because now, a time-challenging thing is open access qualified datasets. So, we don't deal with files that are given to writers upon request. After publishing this study, we will make datasets available to the public, which is helpful in two ways. First, it saves researchers time and money, which lets them put more effort into comparing and evaluating algorithms in a manner that is less biased. We think that this gathering will help researchers choose the best samples for subsequent research. This study also investigates the detection of rice leaf disease used own dataset for the context of Bangladesh's distinctive agricultural terrain. The application of cutting-edge deep learning models, such as light CNN, MobileNet-v2, and InceptionNet-v2, represents a tactical shift toward the use of artificial intelligence for accurate and timely disease diagnosis and also proves these model's performance that datasets are not biased. This study is novel because it directly affects food security and economic results, both of which are significantly impacted by early illness diagnosis with the use of the actual region's own datasets.

The structure of this document is as follows: Section 2 provides an overview of the Literature Review used in this research study. The Datasets Description is proposed in Section 3. Section 4 covers the materials and techniques, including the algorithms and datasets utilized. Section 5 provides a summary of the results obtained from this study. The Conclusion and future work are presented in Section 6.

\section{Literature Review}\label{sec2}

The study of some recent image-processing research works is presented in this section. This review of the literature discusses a number of approaches and strategies for using image processing to detect rice leaf disease. This study considers many algorithmic models to account for the implications of machine learning and deep learning approaches. A few recent research on the identification and classification of diseases affecting rice plants are included in the list below.

Pallapothula Tejaswini et al. \cite{lr1} classified diseases affecting rice leaves. Their suggested 5-layer convolutional network outperforms the other common deep learning models in accuracy by about 6\%. They discovered that they could get a notable level of accuracy with a handcrafted model with fewer layers by modifying the training parameters, such as learning rate, epochs, and optimizer approaches. As they get more adept at identifying Daffodil International University The easier it will be for farmers to safeguard their crops against these six illnesses. With 78.2\% accuracy, the 5-layer convolution model performed the best.Based on the disease stage, some ideas for fertilizers and pesticides are made in this study, which uses an SVM classifier to identify rice ailments. In terms of brown spot, leaf blast, and bacterial leaf blight, this method's accuracy is 90.95 percent, 94.1 percent, and 85.7\%, respectively. Despite being correct, just three diseases were found during this examination.

PNN was used in \cite{lr10} to detect rice illness. This study showed good performance in diagnosing Tungro, Bacterial Leaf Blight, and Brown Spot, with accuracy rates of 92.31\%, 96.25\%, and 97.96\%, respectively. This study did not employ any form characteristics, which might have increased accuracy by aiding in the identification of illnesses. Four machine learning methods have been examined for rice leaf disease detection: KNN, Decision Tree, Logistic Regression, and Naive Bayes. The Decision Tree provides the most accuracy out of all of them.

Pujari et al. \cite{lr22} detected fungal-related diseases in commercial crops as sugarcane, cotton, and chiles. Following their extraction via the discrete wavelet transform (DWT), the characteristics were further reduced using principal wavelet analysis (PWA).

Furthermore, the four primary methods for creating the processing scheme were explained by the study's author \cite{lr11}. The RGB input image is initially given a framework for color transformation. Because HSI is used as a color descriptor, RGB is used to generate color. RGB picture conversion or transportation is also required. The second step involves masking and removing green pixels using a threshold value. Segmenting the picture using the returned useable segments from the previous stage is the third step. After the first stage is completed, segmentation should be finished. 

According to \cite{lr9}, hexagonal matching is utilized for plant disease diagnosis. Since plant diseases mostly impact the leaves, the approach addresses edge detection in addition to color for histogram matching. The training procedure uses layer separation, and the supplied samples are trained. Divide the RGB image's layers into layers representing blue, red, and green to do this. The layered images' edges are likewise created using edge detection technology. It builds the color co-occurrence texture analysis method using spatial gray-level dependency matrices.

The review of the research may be used to deduce that, in an effort to increase productivity, scientists have looked into and used a variety of traditional machine learning-based methods for identifying diseases in rice plants, including k-nearest neighbor (NN) and support vector machine (SVM). In-depth CNN is growing more and more efficient every day, as seen by its higher success rate and more productive outcomes across all classes\cite{sp2}. Even after a long training time, the trained models can distinguish and classify images with ease and speed. The best model available now for large-scale data pattern identification is CNN.

Also, the work \cite{lr12} provides several image-processing techniques for the identification of plant diseases. Research on current methodologies aims to decrease subjectivity in plant disease diagnosis and detection by visual observation.

Zhang et al. \cite{zhang} examined the identification of soybean fungal diseases from a fictitious picture using the study R-CNN. The sustainability of soybeans and the well-being of the agribusiness sector depend on the accurate diagnosis of soybean plant diseases. Despite a number of studies on the subject, the lack of data and technological issues make it more difficult to diagnose illnesses in soybeans. By creating a synthetic soybean plant leaf photo database, this work aims to address the problem of a tiny database. The accuracy percentage obtained from the model's execution was 83.34 percent.

Mohammad et al. \cite{larijani} propose that early detection of rice blast disease may be achieved by image-processing techniques. This study identified the diseases in lab color space using improved k-NN and K-means. The Otsu technique was used for the segmentation of the divided photos. Shape and color features are extracted and used for classification. The k-NN algorithm's and k-means' effectiveness is assessed by taking accuracy, sensitivity, and specificity into account. With this procedure, the accuracy rate was 94\%.

In \cite{lr23} described an innovative method to develop a deep convolutional neural network-based model for plant disease detection. A total of 4500 images were used in the experiment in this investigation. Once minor adjustments were made, the accuracy obtained using the recommended method went from 95.8\% to 96.3\%. Following the model's training on 30880 images, 2589 pictures were used for validation.

Vimal et al.\cite{lrx} proposed a framework for rice disease classification using transfer learning techniques. However, they have only managed to gather 619 images of diseased rice leaves over four different classes, all obtained in the field. They achieved a 91.37\% accuracy rate by using a Support Vector Machine (SVM) for classification and a pre-trained deep convolutional neural network (CNN) for feature extraction. For further extension, they will have an increased dataset for testing the proposed framework and provide a user interface system on actual agriculture areas. 

Krishnamoorthy et al.\cite{lry} proposed a model for rice leaf prediction using deep neural networks with transfer learning. When it had to do with identifying diseases in rice leaf images, they made use of InceptionResNetV2, which is a form of CNN model that was used in conjunction with a transfer learning strategy. By experimenting with various hyperparameters, the basic CNN model was fine-tuned and, after 15 iterations, reached an accuracy of 84.75\%. With the help of 10 epochs and some hyperparameter tuning, InceptionResNetV2 was able to achieve an optimized accuracy of 95.67\%.

The purpose of a literature review is to give an outline of the recent studies and describes that are connected to what they're studying or area of study so that the information can be organized on a piece of paper. To get the best outcome, a balanced dataset is needed. A balanced dataset is required to train and test a dataset using machine learning algorithms. After studying various methods from different researchers, it was found that many experiments did not use an actual field-balanced dataset, which needs to be collected from their region. Furthermore, after analyzing the results of many researchers, it was found that the machine learning algorithms were not used essentially. After implementing various preprocessing techniques, a balanced dataset is established so that the accuracy of the predictive model can improve significantly. If the actual dataset is collected from their region or country and then provides a model or system for appropriately detecting early-stage rice leaf problems, many farmers' agriculture losses ratio can be decreased.

\section{Dataset Descriptions}\label{sec3}

In the following section, we are going to provide a comprehensive description of the dataset that has been proposed.

RiceBD is a pilot dataset that includes images of a rice leaf tree during its whole growth season, which spans most of the time from August to December in Bangladesh. The purpose of this dataset is to construct a representative sample that incorporates the most important cultural features of this plant. Machine learning and deep learning techniques may be used for the classification and identification tasks using this dataset, which is adequate for performance. A total of 1,555 photos were gathered, which were divided into four categories.  A detailed description is shown in Tables 1 and 2, respectively. The reputable agricultural research institute Jamalpur Krishi Gobeshona Institution was also consulted for ideas and recommendations. Their knowledge and verification helped to validate the dataset's validity and applicability, especially concerning disease categorization and identification.

\begin{table}[h]
\caption{ Dataset Description}\label{tab1}%
\begin{tabular}{@{}ll@{}}
\toprule
 RiceBD Datasets &   \\
\midrule
Plant    & Rice Leaf \\
Cultivar   & Oryza sativa \\
Type of data   & RGB Images \\
Data Source Location   & Sylhet \& Dhaka, Bangladesh \\
Annotation    & csv \\
ROI(Region of Interest) captured    & leaf \\
Total size   & 1555 images \\
Application    & The images are appropriate for many machine and deep learning \\
    &  applications, including image identification \& classification. \\
\botrule
\end{tabular}
\end{table}

\begin{table}[h]
\caption{RiceBD Plant is a collection of 1555 images of leaves. The table illustrates the distribution of classes belonging to the leaf images.}\label{tab1}%
\begin{tabular}{@{}lll@{}}
\toprule
 Leaf Symptoms & Severity Levels  & Size\\
\midrule
Healthy    & 0 & 252 \\
Bacterial leaf blight    & 1 & 417  \\
Brown Spot   & 2 & 356 \\
Tungro virus    & 3 & 530  \\

\botrule
\end{tabular}
\end{table}

Images were collected by using various devices, such as a smartphone1 (Galaxy S21 Ultra) and a smartphone2 (Redmi Note 9). As a result, the images have contrasting resolutions, which are 12000 x 6750 and 3000 × 4000, respectively. The configuration of each device is shown in Table 3. For the purpose of data collection, we used two distinct devices since there were a large number of individuals participating, and it was not possible for all of them to have the same technology. Another benefit is that the varied resolutions contribute to the complexity of the dataset, which in turn provides an additional value for the dataset itself. With regard to the fact that it enables diverse and representative inputs to be delivered to the models, the decision to employ various devices is a strategy that is extensively employed in this area of literature. In the real world, agricultural and non-agricultural operators use mobile devices that are distinct from one another in terms of their technological qualities, including their resolution.

\begin{table}[h]
\caption{ Different devices configuration which used image collect}\label{tab1}%
\begin{tabular}{@{}lll@{}}
\toprule
 Attributes & Smartphone1 Camera  & Smartphone2 Camera\\
\midrule
Image size    & 12000 x 6750 & 3000 × 4000 \\
Model device    & Galaxy S21 Ultra & Redmi Note 9  \\
Focal length   & 24mm & 3.5 mm \\
Focal ratio    & f/1.8 & f/2.1  \\
Color space    & RGB & RGB  \\

\botrule
\end{tabular}
\end{table}

The leaves were photographed from an adaxial (upper) angle throughout the course of the entirety of the growing season in a variety of real-life settings, including overcast, sunny, and windy days, with varying degrees of background (other plants and weeds), and levels of background noise. With this acquisition protocol, we were able to achieve several goals, including (i) recording leaves in natural lighting conditions (e.g., (a) indirect sunlight, (b) direct sunlight, (c) strong reflection, and (d) evenly distributed light) (refer to Figure 1), (ii) tracking the development of visual symptoms beyond a period of time, and (iii) capturing leaf as it goes from setting to ripening.

It is well established that mobile pictures have some intrinsic difficulties, such as fluctuating resolution and picture noise, which might impair the precision and clarity of illness diagnosis. To overcome these difficulties, the RiceBD dataset includes photos taken in various settings, which readies machine learning and deep learning models for the complexity and unpredictability of the actual world. In retrospect, the legitimacy of the RiceBD dataset is noteworthy since it has been verified by working with agricultural organizations and carefully capturing actual agricultural circumstances. The dataset provides a thorough and accurate representation of rice leaf pictures thanks to the purposeful use of several devices and diverse conditions, which is crucial for improving classification and identification tasks in agricultural research. For this research on rice leaf disease detection, establish the validity and reliability of the RiceBD dataset by highlighting the validation procedure that involves working with agricultural organizations and purposefully capturing a variety of picture circumstances

\section{MATERIALS AND METHODOLOGY}\label{sec4}

\subsection{Data Acquisition}\label{subsec1}
A procedure that is carried out by persons who are motivated by a variety of goals is known as data collecting, and it is the first stage in any research endeavor. Our agricultural study is focused on the affected rice leaves in our dataset, which serves as the main point of the research we were doing. The aim is to extract valuable insights essential for the advancement of our nation. To achieve this, we embarked on a data collection mission during the prevalent Amon rice season, capitalizing on the extensive rice fields across our country. This endeavor resulted in the capture of over a thousand images encompassing four distinct leaf categories: those portraying manifestations of Leaf Blight, Tungro virus, Brown Spot, and images depicting healthy leaves. All images were meticulously captured using a smartphone, ensuring a comprehensive coverage of angles and perspectives. Despite challenging environmental conditions, including intense heat, the data collection process persevered.

\begin{figure}[h]%
\centering
\includegraphics[width=0.9\textwidth]{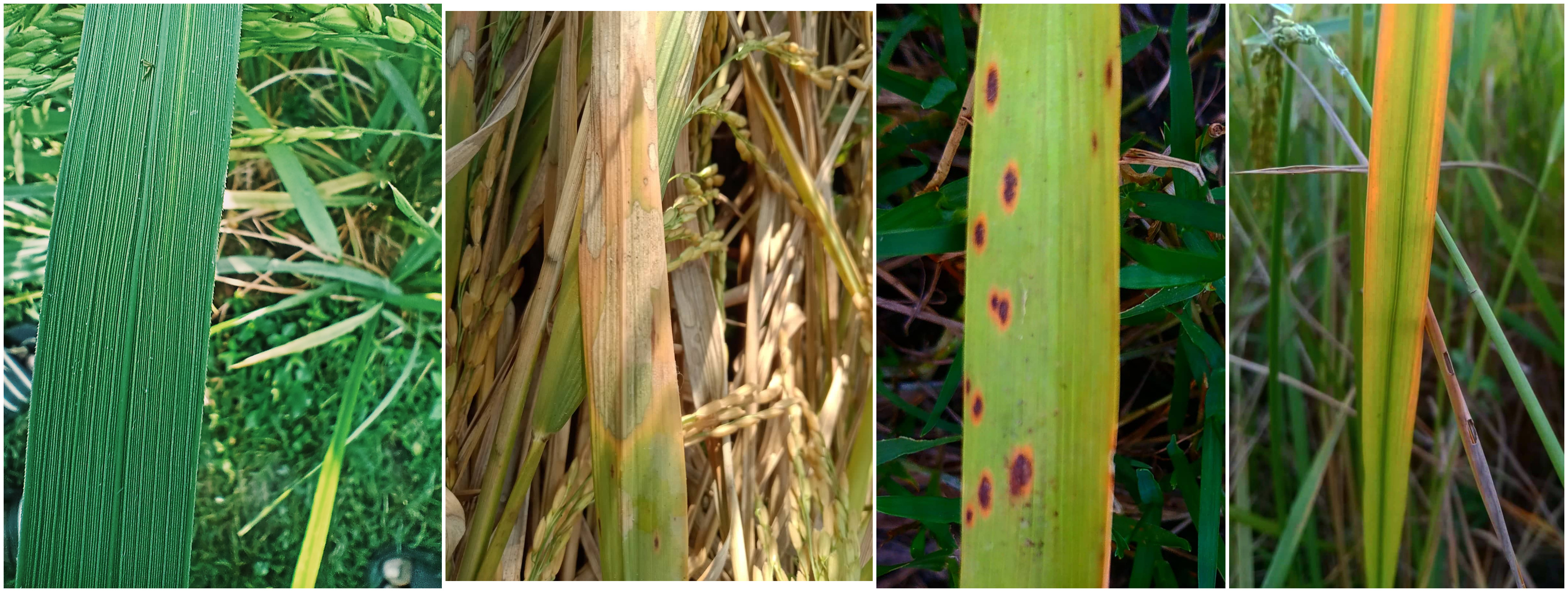}
\caption{Sample images of datasets. (Normal Leaf, Leaf Blight, Brown Spot Leaf, and Tungro virus Leaf are presented from left to right for the above image)}\label{fig1}
\end{figure}

\subsection{Data Preprocessing}\label{subsec2}
Before the implementation phase, this step entails working on the database. It is essential to pre-process the data for each research project before using it for modeling. To utilize the photographs in the right format and facilitate modeling more quickly, normalization of the images is done in this research effort.
\subsubsection*{4.2.1 Image Augmentation}
Image augmentation indicates the process of producing new photographs for the purpose of training our deep learning model throughout the training process. It is not necessary for us to physically collect these new photographs since they are generated by making use of the training images that are currently being made accessible. A crucial component of this study is using the “ImageDataGenerator” class to use image augmentation methods strategically. Several changes are included in the augmentation strategy, such as scaling, rotating, zooming, flipping horizontally and vertically, and shifting. 

\subsubsection*{4.2.2  Data Splitting}
An 80-20 train-test split was instituted in order to conduct a thorough evaluation of the models' capacity for extrapolation. A sufficiently independent test set is maintained for comprehensive assessment according to this meticulous partitioning, which assures that the models are trained on a substantial amount of data.

\subsubsection*{4.2.3  Label Encoding}
Using the ‘LabelEncoder’ function from the scikit-learn package, certain classes were numerically labeled. In the aftermath of this, a critical phase was transforming these labels into one-hot encoding. The training of the models relies substantially on this categorical representation, which correctly discerns and classifies the dataset's varied availability of rice diseases.

\subsubsection*{4.2.4  Data Creation}
The process for creating datasets for both the training and validation stages was methodically carried out by using the ‘flow\_from\_dataframe’ function that is available in TensorFlow. The training dataset was systematically enhanced using augmentations, so exposing the model to a broader range of representations of rice leaf diseases. On the contrary, the validation dataset remained unaltered, assuring that the model's performance could be measured using valid, real-world, and unmodified data.

\begin{figure}[h]%
\centering
\includegraphics[width=0.9\textwidth]{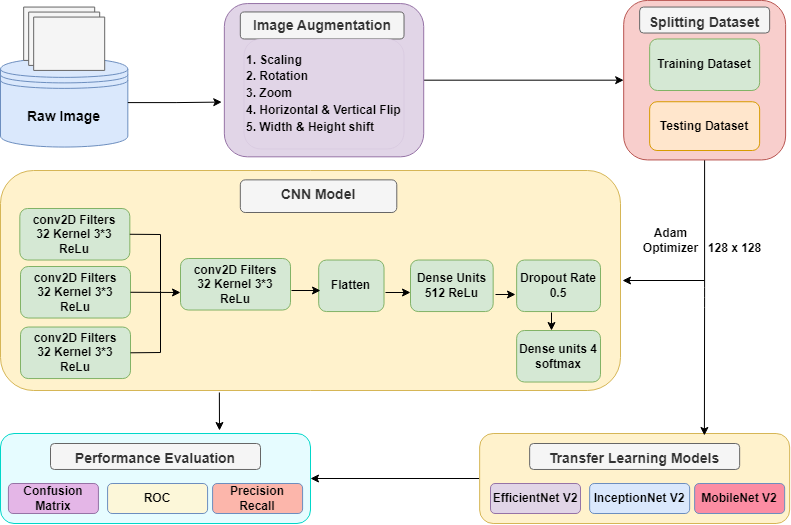}
\caption{Proposed architecture of the Methodology}\label{fig1}
\end{figure}

\subsection{CNN Model}\label{subsec3}
A well-designed convolutional neural network (CNN) for image categorization has been represented by the light CNN\cite{mt3}. This model was constructed using the TensorFlow Keras package and is based on freshly trained weights. The architecture is able to gradually capture the structured features present in input images because it uses max-pooling operations after three convolutional layers. An output layer with a softmax activation function for predicting class probabilities is reached after the flattening layer has linked to densely connected layers that have a dropout mechanism for regularisation\cite{mt4}. In order to conduct dataset analysis, we used five convolutional layers, since this baseline architecture was utilized by a significant number of researchers\cite{mt4-1}. This custom-designed model allows for an extensive evaluation of its image classification ability and serves as a benchmark in our studies, which is illustrated in Figure 3. 
\begin{figure}[h]%
\centering
\includegraphics[width=0.6\textwidth]{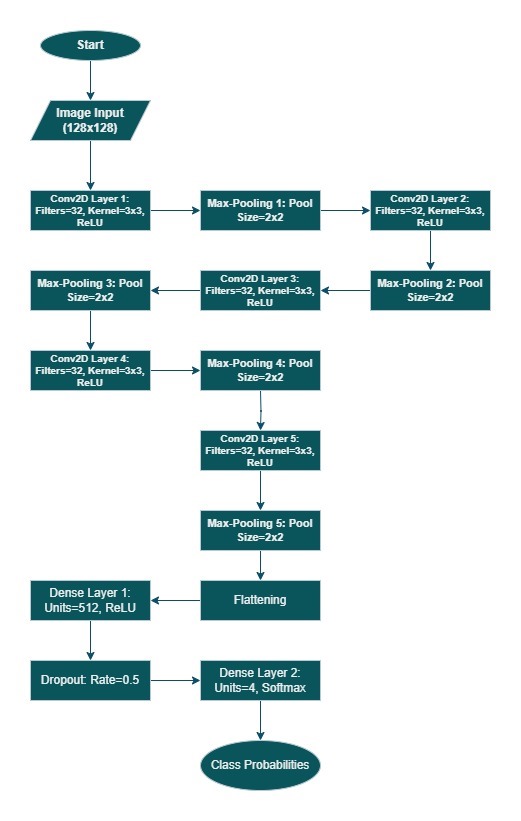}
\caption{Architecture of the CNN}\label{fig1}
\end{figure}

\subsection{Transfer Learning Models}\label{subsec5}
\subsubsection*{4.4.1 InceptionNet-V2}
This is the second version of the DL convolution architecture sequence created by Google \cite{mt5}. The architecture used in image classification is exceptionally refined, using a model that integrates many convolutional filters of varying sizes into a single filter. This approach effectively reduces the number of parameters that need to be learned and the computational complexity associated \cite{mt6}. So that computational complexity is managed without compromising expressive capacity, the model also incorporates factorized convolutions. To achieve an appropriate balance between model complexity and speed, InceptionNetV2 uses architectural enhancements and carefully planned modules, as opposed to EfficientNetV2's combined scaling. InceptionNetV2 is differentiated as inefficient feature extraction due to the inclusion of advanced blocks such as inception and reduction blocks. 

\subsubsection*{4.4.2  MobileNet-V2}
The MobileNet-V2 variant is both more compact and more efficient than prior genres. It applies a single filter to each input channel by using depthwise separable convolutions, and with certain convolutions, it employs a 1x1 filter \cite{mt7}. The separable layers indicate similarities to the convolutional layers in terms of their depth, but they diverge in their approach by performing the filtering and merging processes via the use of two separate layers. When point convolutions are not included, there are a total of 28 layers. With the exception of the fully connected layer that consists of the softmax layer, each layer is succeeded by batch normalization and relu activation layers \cite{mt8}. Using inverted residual blocks with linear bottleneck layers, the architecture finds the best equilibrium between model size and accuracy. Finally, these blocks enable it to be easier to separate features accurately while reducing the number of parameters.

\subsubsection*{4.4.3  EfficientNet-V2}
The EfficientNetV2 convolutional neural network (CNN) models were developed using a compound scaling approach\cite{mt9}. By methodically and synchronously increasing the model's depth, width, and resolution, this methodology improves performance across every dimension\cite{mt10}. Integrating EfficientNetV2 as a modern state-of-the-art model allows us to test its image classification competencies and compare it to other mainstream architectural paradigms, which is helpful in our study.

\begin{figure}[h]%
\centering
\includegraphics[width=1\textwidth]{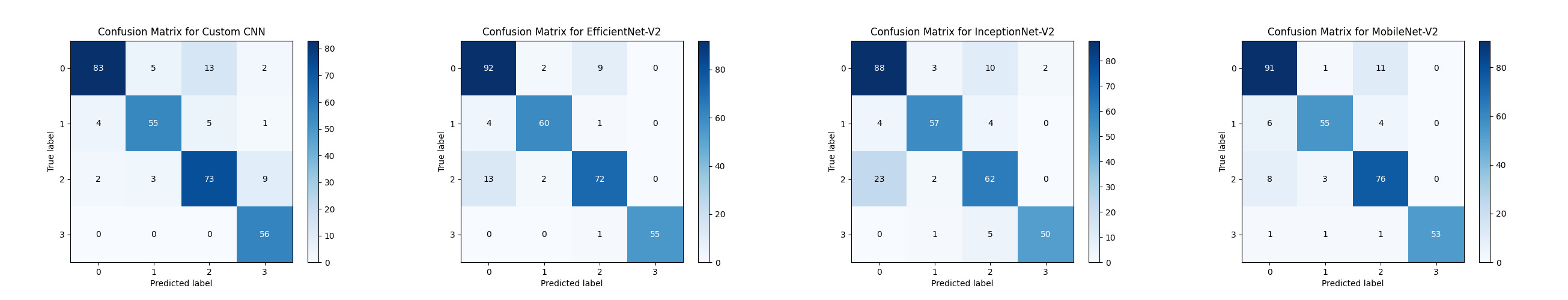}
\caption{Confusion matrix of all models}\label{fig1}
\end{figure}

\subsection{Model Training}\label{subsec5}
These three Transfer Learning models were chosen because many researchers have used them as the base model and suggested which one works best for picture datasets \cite{mt11}. Each transfer learning model was trained with meticulous regard to detail over the course of 50 epochs. The objective of this iterative procedure was to achieve the perfect convergence of the model while simultaneously protecting against potential threats of overfitting this model. Using the augmented training dataset, the training was carried out, and then the validation was performed using a validation dataset that was not adequately augmented. Utilizing Softmax as the activation function and Adam as the optimizer were the two methods that were used. Evaluation measures such as categorical cross-entropy loss and accuracy were used to conduct an in-depth analysis of the performance of the model.

\begin{figure}[h]%
\centering
\includegraphics[width=0.9\textwidth]{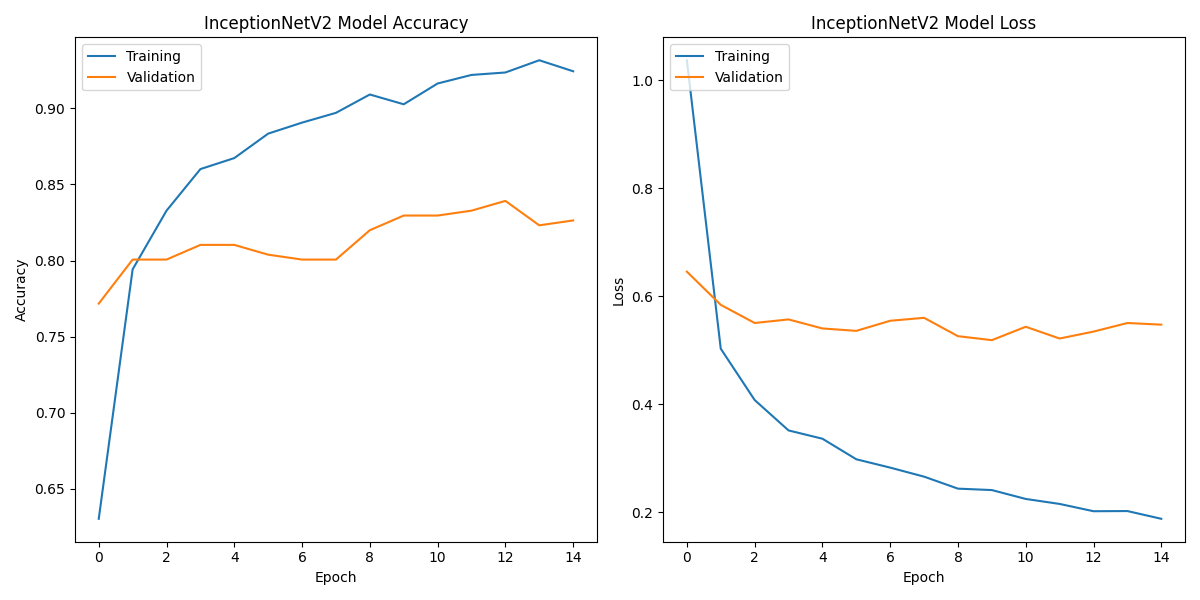}
\caption{Accuracy and loss curve of InceptionNet-V2 model}\label{fig1}
\end{figure}

\begin{figure}[h]%
\centering
\includegraphics[width=0.9\textwidth]{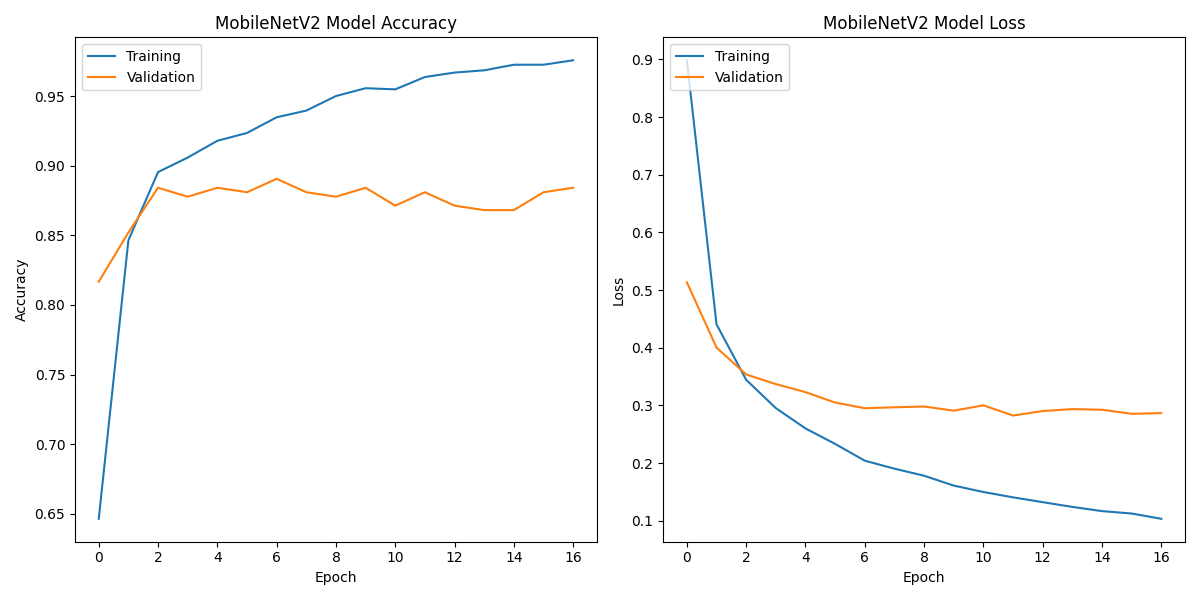}
\caption{Accuracy and loss curve of MobileNet-V2 model}\label{fig1}
\end{figure}

\begin{figure}[h]%
\centering
\includegraphics[width=0.9\textwidth]{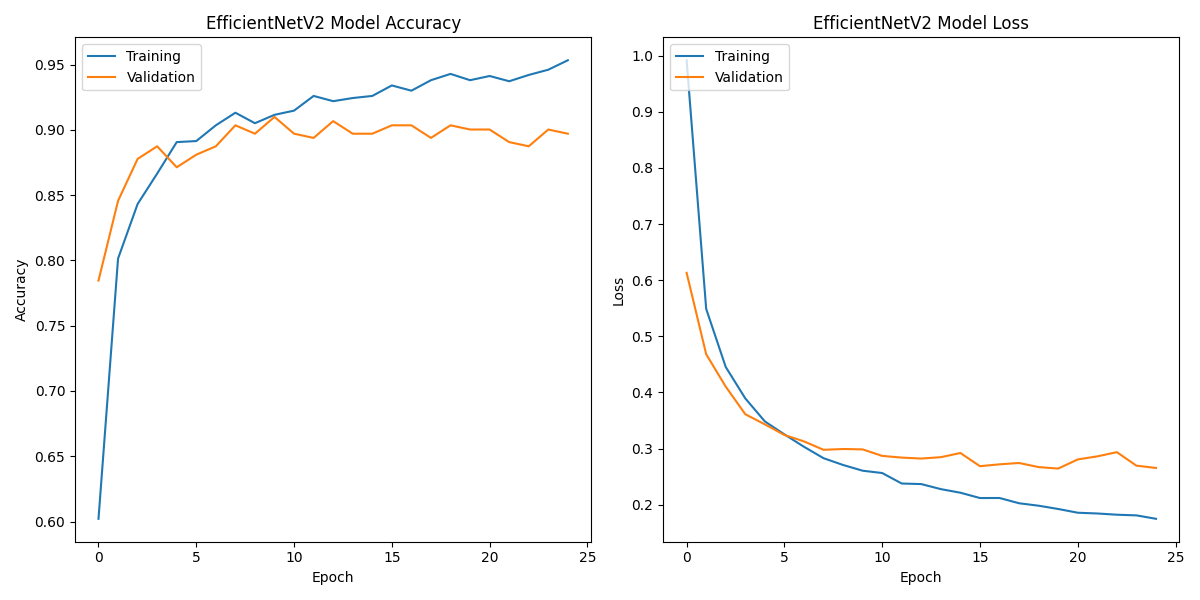}
\caption{Accuracy and loss curve of EfficientNet-V2 model}\label{fig1}
\end{figure}

\begin{figure}[h]%
\centering
\includegraphics[width=0.9\textwidth]{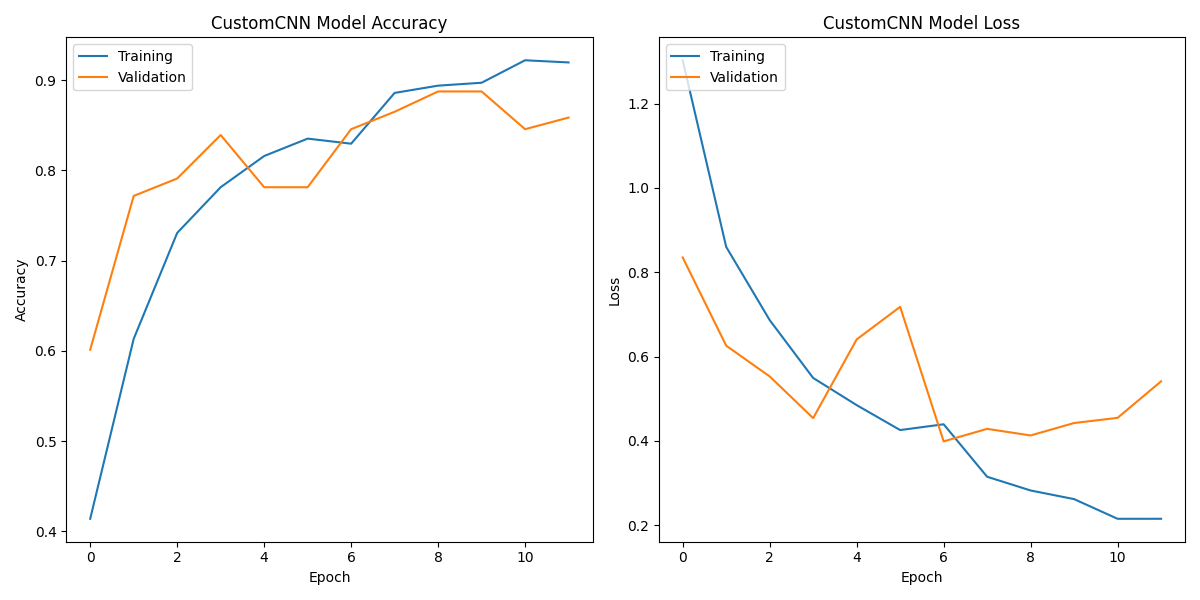}
\caption{Accuracy and loss curve of CNN model}\label{fig1}
\end{figure}

\begin{figure}[h]%
\centering
\includegraphics[width=0.9\textwidth]{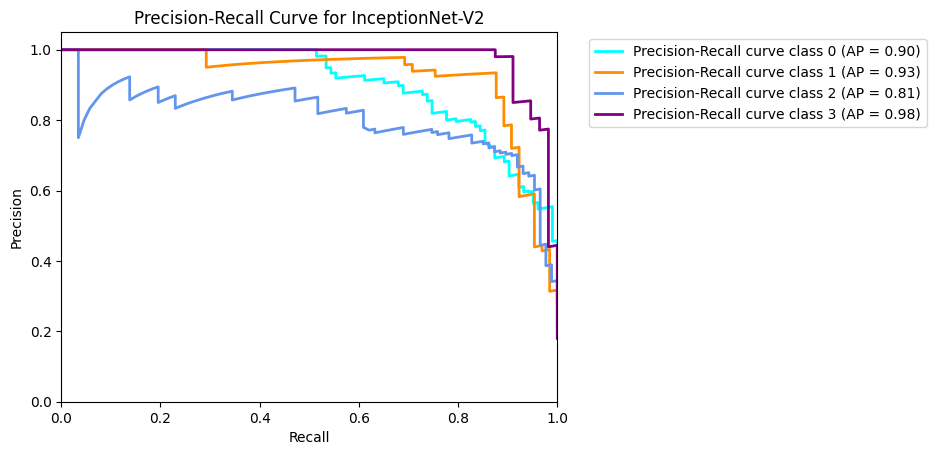}
\caption{Precision-Recall Score of InceptionNet-V2 model}\label{fig1}
\end{figure}

\begin{figure}[h]%
\centering
\includegraphics[width=0.9\textwidth]{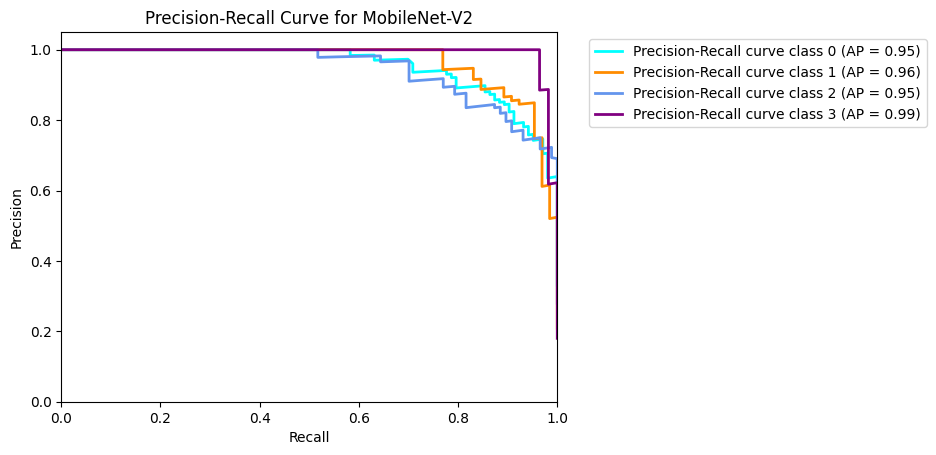}
\caption{Precision-Recall Score curve of MobileNet-V2 model}\label{fig1}
\end{figure}

\begin{figure}[h]%
\centering
\includegraphics[width=0.9\textwidth]{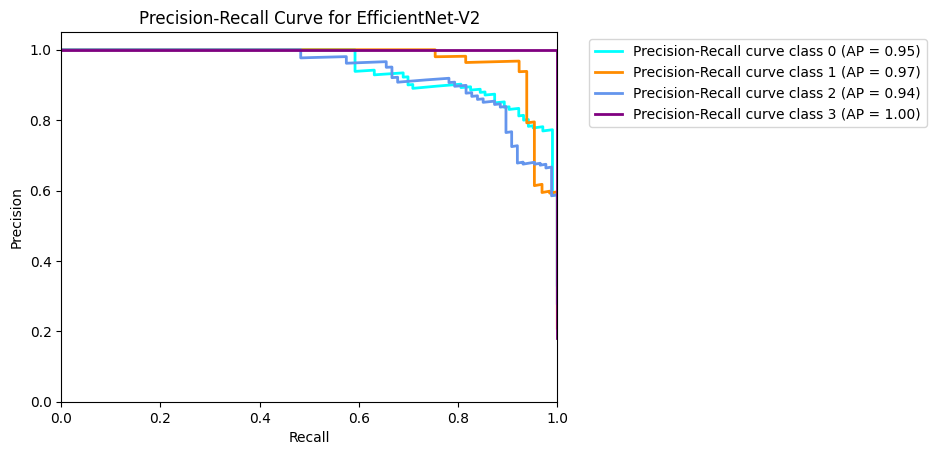}
\caption{Precision-Recall Score curve of EfficientNet-V2 model}\label{fig1}
\end{figure}

\begin{figure}[h]%
\centering
\includegraphics[width=0.9\textwidth]{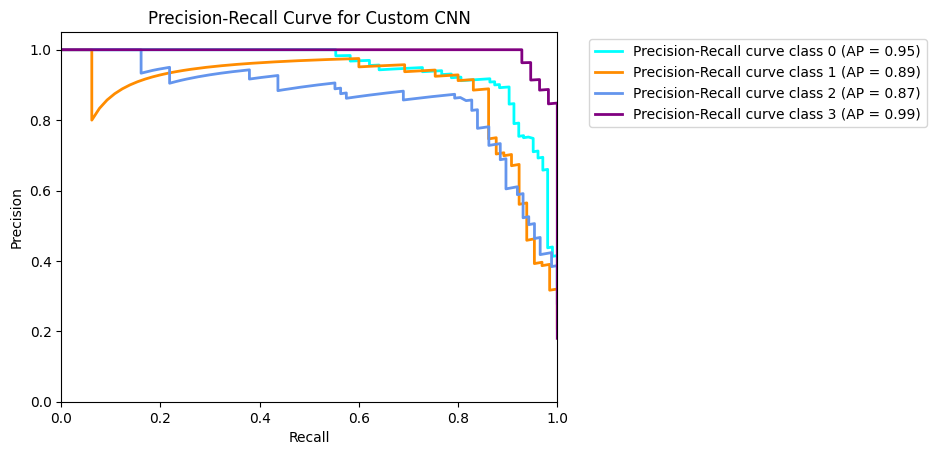}
\caption{Precision-Recall Score of CNN model}\label{fig1}
\end{figure}

\begin{figure}[h]%
\centering
\includegraphics[width=0.9\textwidth]{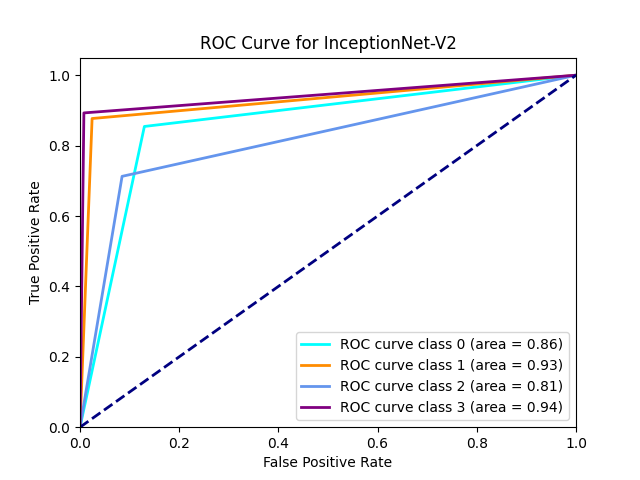}
\caption{ROC curve of InceptionNet-V2 model}\label{fig1}
\end{figure}

\begin{figure}[h]%
\centering
\includegraphics[width=0.9\textwidth]{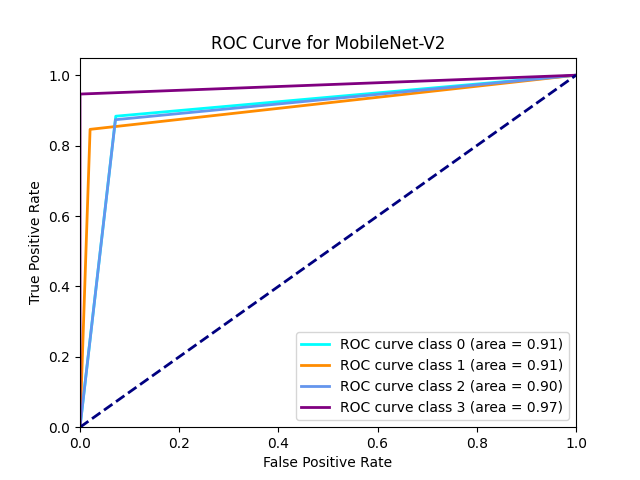}
\caption{ROC curve of MobileNet-V2 model}\label{fig1}
\end{figure}

\begin{figure}[h]%
\centering
\includegraphics[width=0.9\textwidth]{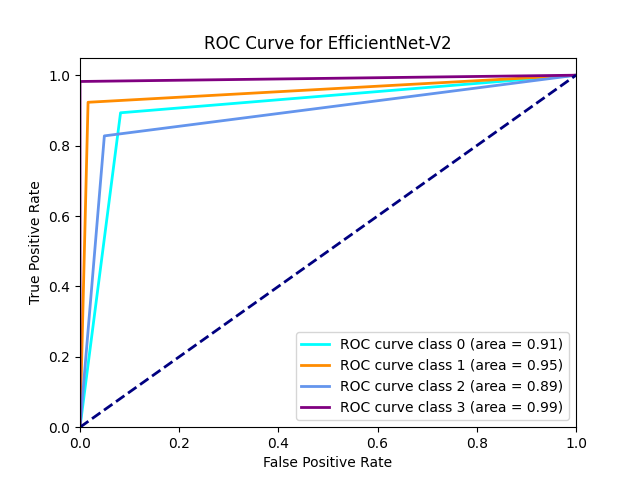}
\caption{ROC curve of EfficientNet-V2 model}\label{fig1}
\end{figure}
\begin{figure}[h]%
\centering
\includegraphics[width=0.9\textwidth]{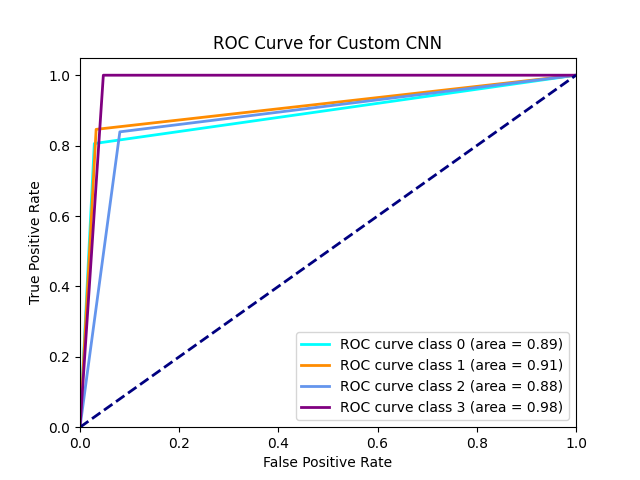}
\caption{ROC curve of CNN model}\label{fig1}
\end{figure}

\section{Result Analysis and Discussion}\label{sec5}

\subsection{Experimental Setup}\label{subsec2}
Experiments were carried out on a system featuring an Intel core processor, 16 GB RAM, and running Windows 10 x64. Google Colaboratory served as the primary platform for conducting simulations and Python programming aimed at identifying rice leaf diseases. Subsequently, a model recommendation surfaced following a thorough analysis of the collected observations.

\subsection{Performance Measures}\label{subsec2}
The investigation compared the performance of pre-trained InceptionNet-V2, MobileNet-V2, and Lite CNN models in identifying and categorizing rice leaf diseases versus healthy leaves using a dataset specific to rice leaf diseases. Our models were configured to process input data in the form of 128x128 images. During training, hyperparameters were optimized using a batch size of 32 over 50 epochs. The dataset's multiclass nature led to the utilization of the softmax activation function in the output layer. As a result, the InceptionNet-V2 model attained 85\% accuracy, while the MobileNet-V2 model reached 89.75\% accuracy, the EfficientNet-V2 model reached 91.5\% accuracy, and lastly, the Light CNN model attained 80.5\% accuracy.

The Rice leaf disease classifier utilizes five performance indicators - accuracy (ACC), precision (PPR), recall or sensitivity (Sen), F1 score, and Area under the ROC curve (AUC) score - tailored for assessing its effectiveness across all analysis datasets.

\begin{table}[h]
\caption{Performance Analysis of InceptionNet-V2}\label{tab1}%
\begin{tabular}{@{}llllll@{}}
\toprule
Model & Classes &  Precision  & Recall & F1-Score & AUC\\
\midrule
InceptionNet-V2 &  Healthy Leaf   & 0.77 & 0.82 & 0.79 & 0.86 \\
    & Bacterial leaf blight & 0.93 & 0.83 & 0.88 & 0.93 \\
   & Brown Spot & 0.76 & 0.78 & 0.77 & 0.81 \\
    & Tungro virus & 0.94 & 0.91 & 0.93 & 0.94  \\

\botrule
\end{tabular}
\end{table}

The InceptionNet-V2 model attained an average precision of 85\%, a recall of 84\%, an F1 score of 84.5\%, and an AUC score of 89\%. Table 4 summarizes the performance metrics for individual classes, providing specific performance indicators for each class within the InceptionNet-V2 model. Also, the confusion matrix, accuracy and loss curve, precision-recall curve, and ROC curve are shown in Figures 3,4,8, and 12, respectively.

Figure: For Models

\begin{table}[h]
\caption{Performance Analysis of MobileNet-V2}\label{tab1}%
\begin{tabular}{@{}llllll@{}}
\toprule
Model & Classes &  Precision  & Recall & F1-Score & AUC\\
\midrule
MovileNet-V2 &  Healthy Leaf   & 0.81 & 0.87 & 0.84 & 0.91 \\
    & Bacterial leaf blight & 0.90 & 0.88 & 0.89 & 0.91 \\
   & Brown Spot & 0.84 & 0.80 & 0.82 & 0.90 \\
    & Tungro virus & 0.98 & 0.95 & 0.96 & 0.97  \\

\botrule
\end{tabular}
\end{table}

The MobileNet-V2 model achieved an overall average accuracy of 89.75\%, with specific performance metrics per class outlined in Table 2. This table-5 provides detailed key performance indicators, such as recall, F1-score, and area under the curve (AUC) score, for individual classes in the model. The recall rate was 89.25\%, the F1 score was 89.5\%, and the AUC score was 92.75\%. The graphs depicting the confusion matrix, accuracy and loss curve, precision-recall curve, and ROC curve are shown in Figures 3,5,9, and 13, respectively.

\begin{table}[h]
\caption{Performance Analysis of CNN model}\label{tab1}%
\begin{tabular}{@{}llllll@{}}
\toprule
Model & Classes &  Precision  & Recall & F1-Score & AUC\\
\midrule
InceptionNet-V2 &  Healthy Leaf   & 0.78 & 0.78 & 0.78 & 0.89 \\
    & Bacterial leaf blight & 0.74 & 0.78 & 0.76 & 0.91 \\
   & Brown Spot & 0.82 & 0.77 & 0.79 & 0.88 \\
    & Tungro virus & 0.88 & 0.91 & 0.89 & 0.98  \\

\botrule
\end{tabular}
\end{table}

\begin{table}[h]
\caption{Performance Analysis of EfficientNet-V2}\label{tab1}%
\begin{tabular}{@{}llllll@{}}
\toprule
Model & Classes &  Precision  & Recall & F1-Score & AUC\\
\midrule
EfficientNet-V2 &  Healthy Leaf   & 0.85 & 0.90 & 0.88 & 0.91 \\
    & Bacterial leaf blight & 0.92 & 0.92 & 0.92 & 0.95 \\
   & Brown Spot & 0.89 & 0.84 & 0.86 & 0.89 \\
    & Tungro virus & 1.0 & 0.98 & 0.99 & 0.99  \\

\botrule
\end{tabular}
\end{table}

The EfficientNet-V2 model achieved an overall average Precision of 91.5\%, with specific performance metrics per class outlined in Table 6. This table provides detailed key performance indicators, such as recall, F1-score, and area under the curve (AUC) score, for individual classes in the model. The recall rate was 91\%, the F1-score was 91.25\%, and the AUC score was 93.5\%.Compare to other model this EfficientNet-V2 achieve the best accuracy. The graphs depicting the confusion matrix, accuracy and loss curve, precision-recall curve, and ROC curve are shown in Figures 3,6,10, and 14, respectively.

The Light CNN model demonstrated an overall average accuracy of 80.5\%, with specific class-wise performance metrics detailed in Table 3. This table-7 presents specific performance indicators, including recall, F1-score, and area under the curve (AUC) score, for each class in the model. The recall rate achieved 80\%, the F1 score was 89.5\%, and the AUC score reached 92.75\%. Graphs illustrating the confusion matrix, accuracy and loss curve, precision-recall curve, and ROC curve are shown in Figures 3,7,11, and 15, respectively.

After assessing and contrasting MobileNet-V2 and EfficientNet-V2 models, it's evident that MobileNet-V2 achieves an average accuracy of 89.75\%, whereas EfficientNet-V2 reaches an accuracy of 91.5\%. This indicates that MobileNet-V2 outperforms InceptionNet-V2 by a notable margin in terms of accuracy. Hence, it's reasonable to conclude that the EfficientNet-V2 model exhibits greater accuracy compared to MobileNet-V2. Based on this performance, we recognized that the RiceLeafBD dataset has a very good and balanced performance, making it a better choice for the study that helps farmers who are negatively impacted.

\subsection{Comparative Aanlysis}\label{subsec2}
Various machine learning and image processing techniques employed in identifying and categorizing diseases in rice are compared and explained in the following text. Table 8 above indicates that our proposed model achieved a commendable accuracy level of 91.5\%, surpassing other models, which means our datasets are not biased datasets, proving the model performance. While M. Akila and P. Deepan's \cite{rc2} model performed well with an 88\% accuracy, it still falls slightly short of our approach. However, it's worth noting that their method requires a substantial amount of time. Shrivastava et al.'s \cite{rc1} approach demonstrated good performance with an accuracy of 91.37\%, but its shortcomings were evident in inferior performance.

These two approaches are closer to our accuracy, although they have notable drawbacks. Other models showed good accuracy, but the significant issues in their methodologies pose challenges. The best thing about our model is that it was trained using a newly collected dataset. Also, when we looked at all of its performances, we found that it performed better than other studies, which indicated that our proposed dataset is essential for other studies that use different analyses and will make a big difference in the world.

\begin{sidewaystable}
\caption{Comparision with other existing works}\label{tab3}
\begin{tabular}{p{2.5cm}p{3.5cm}p{4cm}p{1.5cm}p{3cm}}
\toprule
Authors  & Technique Used  & Disease Identified & Accuracy  & Demerits \\ 
\midrule 
\cite{rc1} & CNN, AlexNet  & RB, BLB, SB, HL& 91.37\% & Challenges in addressing external factors causing noise and lighting issues.  \\ 
\cite{rc2} & R-FNN, RCNN, SSD & Diseases and pests of various plants were identified  & 88\% & Requires a significant amount of time.  \\ 
\cite{rc3} & SVM classifier  & rice blast diseases, narrow brown spot, BLB, brown  & 70\% & Displays the least accuracy in comparison to the others.  \\ 
Proposed Model & InceptionNet-V2,MobileNet-V2,light CNN and EfficientNet-V2 &  There is no demerits  of our model.Also This is cost efficient. & 91.50\%  \\
\bottomrule
\end{tabular}
\footnotetext{}
\footnotetext{}
\end{sidewaystable}

\section{CONCLUSION}\label{sec4}
Identifying and managing rice plant diseases effectively is a big deal for today's farming. This research explores different ways to do this, like using image processing, machine learning, and deep learning. The goal is to help researchers and farmers predict diseases early and act fast. One of the key things this research has done is create a special model that can tell the difference between common rice leaf diseases in Bangladesh. This model is built on Light CNN, InceptionNet-V2, EfficientNet-V2, and MobileNet-V2 and has shown impressive accuracy, with room for even more improvement in the future. Right now, we’re working on making it more reliable and able to handle tough situations like busy backgrounds and different lighting while still keeping it easy to understand. We also proposed our own dataset, which was collected from the Bangladesh field, and elaborately explained our dataset and also gave direction for further research work to serve society using this dataset.

Here, using Light CNN, InceptionNet-V2, EfficientNet-V2, and MobileNet-V2 is a new approach being tried in Bangladesh. It's hard to compare it to other methods because there aren't enough standardized pictures of diseased rice to use as a benchmark. But there's a lot of potential for improvement, primarily if we can collect more pictures of sick rice leaves. Deep learning, especially Convolutional Neural Networks (CNNs), is super crucial for finding diseases in pictures. Our chosen models, including Light CNN, InceptionNet-V2, EfficientNet-V2, and MobileNet-V2, have different levels of accuracy, but they all show that they're efficient and effective and proposed datasets also not biased, which is significant for use research work for Bangladesh farmers. The model we're suggesting is the EfficientNet-V2 model, which is able to correctly classify tungro, brown spot, and blight diseases 90\% of the time when it's trained with 80\% of the pictures and tested with 20\% of them. Using these models shows that they can be really accurate for certain diseases. In conclusion, this review highlights the significance of our specific models and methodologies in rice leaf disease detection. The combination of diverse models and approaches fosters a comprehensive understanding, providing valuable insights for future advancements in precision agriculture. For further work, we intend to expand the dataset, which will significantly contribute to the research community. Additionally, we will apply the Transformer model to improve the accuracy of disease identification, providing valuable assistance to farmers in preventing financial losses.

\backmatter

\bmhead{Acknowledgments}

We express our sincere appreciation to Md. Amin, Mustaqim Ahmed, Ashraful Islam, and Abdus Samad for their invaluable contribution to collecting the rice leaf dataset.

\section*{Declarations}

\begin{itemize}
\item Conflict of interest: The authors declare that they have no conflict of interest.
\item Data Availability: The data supporting the findings of this study will be made available upon request from the authors.
\item Consent for publication: Not applicable.
\item Code availability: The Code supporting the findings of this study will be made available upon request from the authors.

\end{itemize}

\noindent


\bibliography{sn-bibliography}


\begin{thebibliography}{31}
\ifx \bisbn   \undefined \def \bisbn  #1{ISBN #1}\fi
\ifx \binits  \undefined \def \binits#1{#1}\fi
\ifx \bauthor  \undefined \def \bauthor#1{#1}\fi
\ifx \batitle  \undefined \def \batitle#1{#1}\fi
\ifx \bjtitle  \undefined \def \bjtitle#1{#1}\fi
\ifx \bvolume  \undefined \def \bvolume#1{\textbf{#1}}\fi
\ifx \byear  \undefined \def \byear#1{#1}\fi
\ifx \bissue  \undefined \def \bissue#1{#1}\fi
\ifx \bfpage  \undefined \def \bfpage#1{#1}\fi
\ifx \blpage  \undefined \def \blpage #1{#1}\fi
\ifx \burl  \undefined \def \burl#1{\textsf{#1}}\fi
\ifx \doiurl  \undefined \def \doiurl#1{\url{https://doi.org/#1}}\fi
\ifx \betal  \undefined \def \betal{\textit{et al.}}\fi
\ifx \binstitute  \undefined \def \binstitute#1{#1}\fi
\ifx \binstitutionaled  \undefined \def \binstitutionaled#1{#1}\fi
\ifx \bctitle  \undefined \def \bctitle#1{#1}\fi
\ifx \beditor  \undefined \def \beditor#1{#1}\fi
\ifx \bpublisher  \undefined \def \bpublisher#1{#1}\fi
\ifx \bbtitle  \undefined \def \bbtitle#1{#1}\fi
\ifx \bedition  \undefined \def \bedition#1{#1}\fi
\ifx \bseriesno  \undefined \def \bseriesno#1{#1}\fi
\ifx \blocation  \undefined \def \blocation#1{#1}\fi
\ifx \bsertitle  \undefined \def \bsertitle#1{#1}\fi
\ifx \bsnm \undefined \def \bsnm#1{#1}\fi
\ifx \bsuffix \undefined \def \bsuffix#1{#1}\fi
\ifx \bparticle \undefined \def \bparticle#1{#1}\fi
\ifx \barticle \undefined \def \barticle#1{#1}\fi
\bibcommenthead
\ifx \bconfdate \undefined \def \bconfdate #1{#1}\fi
\ifx \botherref \undefined \def \botherref #1{#1}\fi
\ifx \url \undefined \def \url#1{\textsf{#1}}\fi
\ifx \bchapter \undefined \def \bchapter#1{#1}\fi
\ifx \bbook \undefined \def \bbook#1{#1}\fi
\ifx \bcomment \undefined \def \bcomment#1{#1}\fi
\ifx \oauthor \undefined \def \oauthor#1{#1}\fi
\ifx \citeauthoryear \undefined \def \citeauthoryear#1{#1}\fi
\ifx \endbibitem  \undefined \def \endbibitem {}\fi
\ifx \bconflocation  \undefined \def \bconflocation#1{#1}\fi
\ifx \arxivurl  \undefined \def \arxivurl#1{\textsf{#1}}\fi
\csname PreBibitemsHook\endcsname

\bibitem[\protect\citeauthoryear{Islam et~al.}{2022}]{in1}
\begin{barticle}
\bauthor{\bsnm{Islam}, \binits{S.}},
\bauthor{\bsnm{Ghosh}, \binits{S.}},
\bauthor{\bsnm{Podder}, \binits{M.}}:
\batitle{Fifty years of agricultural development in bangladesh: a comparison with india and pakistan}.
\bjtitle{SN Business \& Economics}
\bvolume{2}(\bissue{7}),
\bfpage{71}
(\byear{2022})
\end{barticle}
\endbibitem

\bibitem[\protect\citeauthoryear{Yusrin}{2023}]{in2}
\begin{barticle}
\bauthor{\bsnm{Yusrin}, \binits{N.A.}}:
\batitle{The analysis of rice massive importing in indonesia based on macroeconomics, microeconomics, international economics and politic economics}.
\bjtitle{Ultima Management: Jurnal Ilmu Manajemen}
\bvolume{15}(\bissue{2}),
\bfpage{308}--\blpage{329}
(\byear{2023})
\end{barticle}
\endbibitem

\bibitem[\protect\citeauthoryear{Panayides et~al.}{2020}]{in3}
\begin{barticle}
\bauthor{\bsnm{Panayides}, \binits{A.S.}},
\bauthor{\bsnm{Amini}, \binits{A.}},
\bauthor{\bsnm{Filipovic}, \binits{N.D.}},
\bauthor{\bsnm{Sharma}, \binits{A.}},
\bauthor{\bsnm{Tsaftaris}, \binits{S.A.}},
\bauthor{\bsnm{Young}, \binits{A.}},
\bauthor{\bsnm{Foran}, \binits{D.}},
\bauthor{\bsnm{Do}, \binits{N.}},
\bauthor{\bsnm{Golemati}, \binits{S.}},
\bauthor{\bsnm{Kurc}, \binits{T.}}, \betal:
\batitle{Ai in medical imaging informatics: current challenges and future directions}.
\bjtitle{IEEE journal of biomedical and health informatics}
\bvolume{24}(\bissue{7}),
\bfpage{1837}--\blpage{1857}
(\byear{2020})
\end{barticle}
\endbibitem

\bibitem[\protect\citeauthoryear{Shaikh et~al.}{2022}]{in4}
\begin{barticle}
\bauthor{\bsnm{Shaikh}, \binits{T.A.}},
\bauthor{\bsnm{Rasool}, \binits{T.}},
\bauthor{\bsnm{Lone}, \binits{F.R.}}:
\batitle{Towards leveraging the role of machine learning and artificial intelligence in precision agriculture and smart farming}.
\bjtitle{Computers and Electronics in Agriculture}
\bvolume{198},
\bfpage{107119}
(\byear{2022})
\end{barticle}
\endbibitem

\bibitem[\protect\citeauthoryear{Hughes et~al.}{2015}]{in6}
\begin{botherref}
\oauthor{\bsnm{Hughes}, \binits{D.}},
\oauthor{\bsnm{Salath{\'e}}, \binits{M.}}, et al.:
An open access repository of images on plant health to enable the development of mobile disease diagnostics.
arXiv preprint arXiv:1511.08060
(2015)
\end{botherref}
\endbibitem

\bibitem[\protect\citeauthoryear{Barbedo}{2019}]{in7}
\begin{barticle}
\bauthor{\bsnm{Barbedo}, \binits{J.G.A.}}:
\batitle{Plant disease identification from individual lesions and spots using deep learning}.
\bjtitle{Biosystems engineering}
\bvolume{180},
\bfpage{96}--\blpage{107}
(\byear{2019})
\end{barticle}
\endbibitem

\bibitem[\protect\citeauthoryear{Tejaswini et~al.}{2022}]{lr1}
\begin{bchapter}
\bauthor{\bsnm{Tejaswini}, \binits{P.}},
\bauthor{\bsnm{Singh}, \binits{P.}},
\bauthor{\bsnm{Ramchandani}, \binits{M.}},
\bauthor{\bsnm{Rathore}, \binits{Y.K.}},
\bauthor{\bsnm{Janghel}, \binits{R.R.}}:
\bctitle{Rice leaf disease classification using cnn}.
In: \bbtitle{IOP Conference Series: Earth and Environmental Science},
vol. \bseriesno{1032},
p. \bfpage{012017}
(\byear{2022}).
\bcomment{IOP Publishing}
\end{bchapter}
\endbibitem

\bibitem[\protect\citeauthoryear{Islam et~al.}{2021}]{lr10}
\begin{barticle}
\bauthor{\bsnm{Islam}, \binits{A.}},
\bauthor{\bsnm{Islam}, \binits{R.}},
\bauthor{\bsnm{Haque}, \binits{S.}},
\bauthor{\bsnm{Islam}, \binits{S.}},
\bauthor{\bsnm{Khan}, \binits{M.A.I.}}:
\batitle{Rice leaf disease recognition using local threshold based segmentation and deep cnn}.
\bjtitle{Int. J. Intell. Syst. Appl}
\bvolume{13}(\bissue{5}),
\bfpage{35}--\blpage{45}
(\byear{2021})
\end{barticle}
\endbibitem

\bibitem[\protect\citeauthoryear{Pujari et~al.}{2013}]{lr22}
\begin{barticle}
\bauthor{\bsnm{Pujari}, \binits{J.D.}},
\bauthor{\bsnm{Yakkundimath}, \binits{R.}},
\bauthor{\bsnm{Byadgi}, \binits{A.S.}}:
\batitle{Automatic fungal disease detection based on wavelet feature extraction and pca analysis in commercial crops}.
\bjtitle{International Journal of Image, Graphics and Signal Processing}
\bvolume{6}(\bissue{1}),
\bfpage{24}--\blpage{31}
(\byear{2013})
\end{barticle}
\endbibitem

\bibitem[\protect\citeauthoryear{Sethy et~al.}{2020}]{lr11}
\begin{barticle}
\bauthor{\bsnm{Sethy}, \binits{P.K.}},
\bauthor{\bsnm{Barpanda}, \binits{N.K.}},
\bauthor{\bsnm{Rath}, \binits{A.K.}},
\bauthor{\bsnm{Behera}, \binits{S.K.}}:
\batitle{Deep feature based rice leaf disease identification using support vector machine}.
\bjtitle{Computers and Electronics in Agriculture}
\bvolume{175},
\bfpage{105527}
(\byear{2020})
\end{barticle}
\endbibitem

\bibitem[\protect\citeauthoryear{Chouhan et~al.}{2021}]{lr9}
\begin{barticle}
\bauthor{\bsnm{Chouhan}, \binits{S.S.}},
\bauthor{\bsnm{Singh}, \binits{U.P.}},
\bauthor{\bsnm{Sharma}, \binits{U.}},
\bauthor{\bsnm{Jain}, \binits{S.}}:
\batitle{Leaf disease segmentation and classification of jatropha curcas l. and pongamia pinnata l. biofuel plants using computer vision based approaches}.
\bjtitle{Measurement}
\bvolume{171},
\bfpage{108796}
(\byear{2021})
\end{barticle}
\endbibitem

\bibitem[\protect\citeauthoryear{Chowdhury et~al.}{2023}]{sp2}
\begin{barticle}
\bauthor{\bsnm{Chowdhury}, \binits{M.J.U.}},
\bauthor{\bsnm{Mou}, \binits{Z.I.}},
\bauthor{\bsnm{Afrin}, \binits{R.}},
\bauthor{\bsnm{Kibria}, \binits{S.}}:
\batitle{Plant leaf disease detection and classification using deep learning: A review and a proposed system on bangladesh{\^a}€™ s perspective}.
\bjtitle{International Journal of Science and Business}
\bvolume{28}(\bissue{1}),
\bfpage{193}--\blpage{204}
(\byear{2023})
\end{barticle}
\endbibitem

\bibitem[\protect\citeauthoryear{Yadav et~al.}{2021}]{lr12}
\begin{barticle}
\bauthor{\bsnm{Yadav}, \binits{S.}},
\bauthor{\bsnm{Sengar}, \binits{N.}},
\bauthor{\bsnm{Singh}, \binits{A.}},
\bauthor{\bsnm{Singh}, \binits{A.}},
\bauthor{\bsnm{Dutta}, \binits{M.K.}}:
\batitle{Identification of disease using deep learning and evaluation of bacteriosis in peach leaf}.
\bjtitle{Ecological Informatics}
\bvolume{61},
\bfpage{101247}
(\byear{2021})
\end{barticle}
\endbibitem

\bibitem[\protect\citeauthoryear{Zhang et~al.}{2021}]{zhang}
\begin{barticle}
\bauthor{\bsnm{Zhang}, \binits{K.}},
\bauthor{\bsnm{Wu}, \binits{Q.}},
\bauthor{\bsnm{Chen}, \binits{Y.}}:
\batitle{Detecting soybean leaf disease from synthetic image using multi-feature fusion faster r-cnn}.
\bjtitle{Computers and Electronics in Agriculture}
\bvolume{183},
\bfpage{106064}
(\byear{2021})
\end{barticle}
\endbibitem

\bibitem[\protect\citeauthoryear{Larijani et~al.}{2019}]{larijani}
\begin{barticle}
\bauthor{\bsnm{Larijani}, \binits{M.R.}},
\bauthor{\bsnm{Asli-Ardeh}, \binits{E.A.}},
\bauthor{\bsnm{Kozegar}, \binits{E.}},
\bauthor{\bsnm{Loni}, \binits{R.}}:
\batitle{Evaluation of image processing technique in identifying rice blast disease in field conditions based on knn algorithm improvement by k-means}.
\bjtitle{Food science \& nutrition}
\bvolume{7}(\bissue{12}),
\bfpage{3922}--\blpage{3930}
(\byear{2019})
\end{barticle}
\endbibitem

\bibitem[\protect\citeauthoryear{Guan}{2021}]{lr23}
\begin{bchapter}
\bauthor{\bsnm{Guan}, \binits{X.}}:
\bctitle{A novel method of plant leaf disease detection based on deep learning and convolutional neural network}.
In: \bbtitle{2021 6th International Conference on Intelligent Computing and Signal Processing (ICSP)},
pp. \bfpage{816}--\blpage{819}
(\byear{2021}).
\bcomment{IEEE}
\end{bchapter}
\endbibitem

\bibitem[\protect\citeauthoryear{Shrivastava et~al.}{2019}]{lrx}
\begin{barticle}
\bauthor{\bsnm{Shrivastava}, \binits{V.K.}},
\bauthor{\bsnm{Pradhan}, \binits{M.K.}},
\bauthor{\bsnm{Minz}, \binits{S.}},
\bauthor{\bsnm{Thakur}, \binits{M.P.}}:
\batitle{Rice plant disease classification using transfer learning of deep convolution neural network}.
\bjtitle{The International Archives of the Photogrammetry, Remote Sensing and Spatial Information Sciences}
\bvolume{42},
\bfpage{631}--\blpage{635}
(\byear{2019})
\end{barticle}
\endbibitem

\bibitem[\protect\citeauthoryear{Krishnamoorthy et~al.}{2021}]{lry}
\begin{barticle}
\bauthor{\bsnm{Krishnamoorthy}, \binits{N.}},
\bauthor{\bsnm{Prasad}, \binits{L.N.}},
\bauthor{\bsnm{Kumar}, \binits{C.P.}},
\bauthor{\bsnm{Subedi}, \binits{B.}},
\bauthor{\bsnm{Abraha}, \binits{H.B.}},
\bauthor{\bsnm{Sathishkumar}, \binits{V.}}:
\batitle{Rice leaf diseases prediction using deep neural networks with transfer learning}.
\bjtitle{Environmental Research}
\bvolume{198},
\bfpage{111275}
(\byear{2021})
\end{barticle}
\endbibitem

\bibitem[\protect\citeauthoryear{Zhong et~al.}{2019}]{mt3}
\begin{barticle}
\bauthor{\bsnm{Zhong}, \binits{B.}},
\bauthor{\bsnm{Xing}, \binits{X.}},
\bauthor{\bsnm{Love}, \binits{P.}},
\bauthor{\bsnm{Wang}, \binits{X.}},
\bauthor{\bsnm{Luo}, \binits{H.}}:
\batitle{Convolutional neural network: Deep learning-based classification of building quality problems}.
\bjtitle{Advanced Engineering Informatics}
\bvolume{40},
\bfpage{46}--\blpage{57}
(\byear{2019})
\end{barticle}
\endbibitem

\bibitem[\protect\citeauthoryear{Jeczmionek and Kowalski}{2021}]{mt4}
\begin{barticle}
\bauthor{\bsnm{Jeczmionek}, \binits{E.}},
\bauthor{\bsnm{Kowalski}, \binits{P.A.}}:
\batitle{Flattening layer pruning in convolutional neural networks}.
\bjtitle{Symmetry}
\bvolume{13}(\bissue{7}),
\bfpage{1147}
(\byear{2021})
\end{barticle}
\endbibitem

\bibitem[\protect\citeauthoryear{Albawi et~al.}{2017}]{mt4-1}
\begin{bchapter}
\bauthor{\bsnm{Albawi}, \binits{S.}},
\bauthor{\bsnm{Mohammed}, \binits{T.A.}},
\bauthor{\bsnm{Al-Zawi}, \binits{S.}}:
\bctitle{Understanding of a convolutional neural network}.
In: \bbtitle{2017 International Conference on Engineering and Technology (ICET)},
pp. \bfpage{1}--\blpage{6}
(\byear{2017}).
\bcomment{Ieee}
\end{bchapter}
\endbibitem

\bibitem[\protect\citeauthoryear{Szegedy et~al.}{2016}]{mt5}
\begin{bchapter}
\bauthor{\bsnm{Szegedy}, \binits{C.}},
\bauthor{\bsnm{Vanhoucke}, \binits{V.}},
\bauthor{\bsnm{Ioffe}, \binits{S.}},
\bauthor{\bsnm{Shlens}, \binits{J.}},
\bauthor{\bsnm{Wojna}, \binits{Z.}}:
\bctitle{Rethinking the inception architecture for computer vision}.
In: \bbtitle{Proceedings of the IEEE Conference on Computer Vision and Pattern Recognition},
pp. \bfpage{2818}--\blpage{2826}
(\byear{2016})
\end{bchapter}
\endbibitem

\bibitem[\protect\citeauthoryear{Ucar}{2021}]{mt6}
\begin{barticle}
\bauthor{\bsnm{Ucar}, \binits{M.}}:
\batitle{Diagnosis of glaucoma disease using convolutional neural network architectures}.
\bjtitle{Dokuz Eylul Uni. Fac. of Eng. J. of Sci. and Eng}
\bvolume{23}(\bissue{68}),
\bfpage{521}--\blpage{529}
(\byear{2021})
\end{barticle}
\endbibitem

\bibitem[\protect\citeauthoryear{Howard et~al.}{2017}]{mt7}
\begin{botherref}
\oauthor{\bsnm{Howard}, \binits{A.G.}},
\oauthor{\bsnm{Zhu}, \binits{M.}},
\oauthor{\bsnm{Chen}, \binits{B.}},
\oauthor{\bsnm{Kalenichenko}, \binits{D.}},
\oauthor{\bsnm{Wang}, \binits{W.}},
\oauthor{\bsnm{Weyand}, \binits{T.}},
\oauthor{\bsnm{Andreetto}, \binits{M.}},
\oauthor{\bsnm{Adam}, \binits{H.}}:
Mobilenets: Efficient convolutional neural networks for mobile vision applications.
arXiv preprint arXiv:1704.04861
(2017)
\end{botherref}
\endbibitem

\bibitem[\protect\citeauthoryear{Zeren et~al.}{2020}]{mt8}
\begin{botherref}
\oauthor{\bsnm{Zeren}, \binits{M.T.}},
\oauthor{\bsnm{Aytulun}, \binits{S.K.}},
\oauthor{\bsnm{KIRELL{\.I}}, \binits{Y.}}:
Comparison of ssd and faster r-cnn algorithms to detect the airports with data set which obtained from unmanned aerial vehicles and satellite images.
Avrupa Bilim ve Teknoloji Dergisi
(19),
643--658
(2020)
\end{botherref}
\endbibitem

\bibitem[\protect\citeauthoryear{Nguyen et~al.}{2022}]{mt9}
\begin{bchapter}
\bauthor{\bsnm{Nguyen}, \binits{H.T.}},
\bauthor{\bsnm{Le}, \binits{L.N.}},
\bauthor{\bsnm{Vo}, \binits{T.M.}},
\bauthor{\bsnm{Pham}, \binits{D.N.T.}},
\bauthor{\bsnm{Tran}, \binits{D.T.}}:
\bctitle{Breast ultrasound image classification using efficientnetv2 and shallow neural network architectures}.
In: \bbtitle{Computational Intelligence in Security for Information Systems Conference},
pp. \bfpage{130}--\blpage{142}
(\byear{2022}).
\bcomment{Springer}
\end{bchapter}
\endbibitem

\bibitem[\protect\citeauthoryear{Tan and Le}{2021}]{mt10}
\begin{bchapter}
\bauthor{\bsnm{Tan}, \binits{M.}},
\bauthor{\bsnm{Le}, \binits{Q.}}:
\bctitle{Efficientnetv2: Smaller models and faster training}.
In: \bbtitle{International Conference on Machine Learning},
pp. \bfpage{10096}--\blpage{10106}
(\byear{2021}).
\bcomment{PMLR}
\end{bchapter}
\endbibitem

\bibitem[\protect\citeauthoryear{Weiss et~al.}{2016}]{mt11}
\begin{barticle}
\bauthor{\bsnm{Weiss}, \binits{K.}},
\bauthor{\bsnm{Khoshgoftaar}, \binits{T.M.}},
\bauthor{\bsnm{Wang}, \binits{D.}}:
\batitle{A survey of transfer learning}.
\bjtitle{Journal of Big data}
\bvolume{3}(\bissue{1}),
\bfpage{1}--\blpage{40}
(\byear{2016})
\end{barticle}
\endbibitem

\bibitem[\protect\citeauthoryear{Akila and Deepan}{2018}]{rc2}
\begin{barticle}
\bauthor{\bsnm{Akila}, \binits{M.}},
\bauthor{\bsnm{Deepan}, \binits{P.}}:
\batitle{Detection and classification of plant leaf diseases by using deep learning algorithm}.
\bjtitle{International Journal of Engineering Research \& Technology (IJERT)}
\bvolume{6}(\bissue{7}),
\bfpage{1}--\blpage{5}
(\byear{2018})
\end{barticle}
\endbibitem

\bibitem[\protect\citeauthoryear{Shrivastava et~al.}{2019}]{rc1}
\begin{barticle}
\bauthor{\bsnm{Shrivastava}, \binits{V.K.}},
\bauthor{\bsnm{Pradhan}, \binits{M.K.}},
\bauthor{\bsnm{Minz}, \binits{S.}},
\bauthor{\bsnm{Thakur}, \binits{M.P.}}:
\batitle{Rice plant disease classification using transfer learning of deep convolution neural network}.
\bjtitle{The International Archives of the Photogrammetry, Remote Sensing and Spatial Information Sciences}
\bvolume{42},
\bfpage{631}--\blpage{635}
(\byear{2019})
\end{barticle}
\endbibitem

\bibitem[\protect\citeauthoryear{Suman and Dhruvakumar}{2015}]{rc3}
\begin{barticle}
\bauthor{\bsnm{Suman}, \binits{T.}},
\bauthor{\bsnm{Dhruvakumar}, \binits{T.}}:
\batitle{Classification of paddy leaf diseases using shape and color features}.
\bjtitle{IJEEE}
\bvolume{7}(\bissue{01}),
\bfpage{239}--\blpage{250}
(\byear{2015})
\end{barticle}
\endbibitem

\end{thebibliography}

\end{document}